\documentclass[sigconf,authorversion,nonacm]{acmart}

\usepackage{url}
\usepackage{graphicx}
\usepackage{float}
\usepackage{enumitem}
\usepackage{xcolor}
\usepackage{fancyhdr}
\usepackage{amsmath}
\usepackage{sidecap}
\usepackage{etoolbox}
\usepackage{verbatim}
\usepackage{caption}
\usepackage{amsmath}
\usepackage{subfigure}

\usepackage[linesnumbered,ruled,boxed,noline,noend]{algorithm2e}

\hypersetup{
    colorlinks=true,
    anchorcolor=blue,
    linkcolor=black,
    citecolor=black,
    urlcolor=black,
}

\newcommand{\header}[1]{\vspace{1mm}\noindent{\bfseries{#1.}}}

\begin{document}

\title{Towards Loosely-Coupling Knowledge Graph Embeddings and Ontology-based Reasoning}

\author{Zoi Kaoudi}
\affiliation{%
  \institution{Technische Universit\"{a}t Berlin}
  \country{Germany}
}
%\email{zoi.kaoudi@tu-berlin.de}

\author{Abelardo Carlos Martinez Lorenzo*}\thanks{*Work done while conducting his master thesis at TU Berlin.}
\affiliation{%
	\institution{Sapienza University of Rome}
	\country{Italy}
}

\author{Volker Markl}
\affiliation{%
	\institution{Technische Universit\"{a}t Berlin}
	\country{Germany}
}
%\email{volker.markl@tu-berlin.de}

\begin{abstract}
Knowledge graph completion (a.k.a.~link prediction), i.e.,~the task of inferring missing information from knowledge graphs, is a widely used task in many applications, such as product recommendation 	and question answering. The state-of-the-art approaches of knowledge graph embeddings and/or rule mining and reasoning are data-driven and, thus, solely based on the information the input knowledge graph contains. This leads to unsatisfactory prediction results which make such solutions inapplicable to crucial domains such as healthcare. 
To further enhance the accuracy of knowledge graph completion we propose to loosely-couple the data-driven power of knowledge graph embeddings with domain-specific reasoning stemming from experts or entailment regimes (e.g., OWL2). 
In this way, we not only enhance the prediction accuracy with domain knowledge that may not be included in the input knowledge graph but also allow users to plugin their own knowledge graph embedding and reasoning method.
Our initial results show that we enhance the MRR accuracy of vanilla knowledge graph embeddings by up to 3x and outperform hybrid solutions that combine knowledge graph embeddings with rule mining and reasoning up to 3.5x MRR.
\end{abstract}

\maketitle

\section{Introduction}
\label{sec:intro}
%!TEX root = main.tex

Knowledge graphs (KGs) are extensively used in many application domains, such as search engines, product recommendation, and bioinformatics. 
%A KG consists of a set of triples of the form $(s, r, o)$, which denote that a node $s$ (subject) is connected to a node $o$ (object) via a relationship $r$. 
Nodes in the graph denote entities and edges represent the relations between entities. 
KGs are very often accompanied by ontologies which specify the schema that a KG follows, i.e.,~the type of entities and the relationships between different types. 
KGs are typically stored in data management systems that allow for efficient retrieval of parts of the graphs based on a declarative query~\cite{rdf3x,cliquesquare,kaoudi-survey,cliquesquare-demo}. 
As KGs are usually created from incomplete data sources, it is often the case that there is missing information in the results of such queries. 
The task of \textit{KG completion} (a.k.a.~\textit{link prediction}) aims to infer such missing information. 

Two known methods used for the problem of KG completion is knowledge graph embeddings (KGE) and rule mining.
KGEs are representations of entities and relations into a low-dimensional space which can be used to predict whether a missing link in the graph is true or not.
Rule-mining methods mine logical rules from the KG and then apply them to reason about new information.
Unfortunately, both methods fall short in providing good prediction results for any arbitrary KG.
The reason for this is that they both rely on the data and patterns existing in the input KG and are thus unable to generalize in cases where there is inadequate information. 
For instance, KGEs work well in predicting links for popular entities and relations, i.e.,~dense parts of the graph~\cite{uai-paper20}.
Similarly, rule-mining methods work well when there is enough evidence supporting the patterns they mine.
However, real-world KGs consist of both dense and sparse areas and thus, current approaches fail to provide satisfactory prediction results in the general case and highly depend on the input data.

%mapping entities and relations into a low-dimensional space and using this representation to predict whether a triple is true or not.
%The computed embeddings can then be used to predict missing information from a KG.  
%KGEs are used to answer queries of the form $(?, r, o)$ and $(s, r, ?)$ or whether a triple is true or not.
%However, it is well-known that KGE-based proposals are unable to predict missing information for sparse entities and relations, i.e.,~entities and relations for which there is not much information in the KG~\cite{uai-paper20}. 
%An alternative direction for predicting missing links is rule-based methods which mine logical rules based on the patterns existing in the KG. 
%These rules are then applied to existing triples in the KG to infer new information. 
%However, similarly to KGE-based approaches, these rule-based methods also fall short in providing good results when there is not much evidence to extract rules.
%%Rule-based methods are better suited to provide explainability to the inferred triples.

%In an effort to combine the  the drawbacks of the two aforementioned KG completion directions, 
Recently there have been few proposals on hybrid solutions, combining KGEs with rule mining and reasoning~\cite{kale,plogicnet}.
%These hybrid proposals view the two directions complementary instead of competitors~\cite{rainer-iswc18}. 
%Combining the two approaches is not straightforward. One approach, where KGEs subsume reasoning, is to plug rules in the embedding algorithms themselves by for example adapting the loss function. Another approach, where reasoning subsumes KGEs, is to create rules out of the embeddings and use these rules to infer new information. A third approach is to use both methods simultaneously. In the first two approaches, it is challenging to come up with a common ground based on which KGEs and reasoning can work. In the third approach, it is challenging to determine how the modules should be communicating so that training time does not get prohibitively high.
%Recently there have been few efforts on combining KGEs with reasoning in different ways. 
%In~\cite{kale} the authors propose to jointly embed triples and logical rules in a mathematical framework that uses a common loss function. In contrast, the authors of~\cite{plogicnet} aim at improving Markov Logic Networks, a rule-based approach that combines first-order logic with probabilistic graphical models, with the use of KG embeddings. 
%IterE~\cite{iterE} is a framework that iteratively learns embeddings and rules, and uses the rules to produce new triples that can further improve the embeddings. 
Although they achieve slightly better predictive performance than the monolithic approaches, they still fail to solve the problem as both their embedding and rule-mining components are data-driven. % they rely on evidence that exists in the KG. 
This means that in lack of sufficient evidence in the KG such hybrid methods are also unable to infer useful information. 
In addition, these solutions, are tightly coupled with specific KGE model and rule-based reasoning methods and require significant effort to adapt them to new, more promising ones, if possible at all. 
%IterE is a more flexible but still requires KGE methods that follow the linear map assumption (i.e., the subject embedding can be linearly mapped via the relations embeddings to the object embeddings), such as ANALOGY~\cite{analogy}. 

A natural remedy to improve KG prediction performance is the inclusion of domain expertise.
%For example, \zoi{give an example with healthcare or bioinformatics}
Domain knowledge can be encoded by ontologies and rules which are either user-defined or specified by an entailment regime (e.g.,~RDFS or OWL2).
However, none of the aforementioned approaches leverages such information.
In fact, \cite{GBS-KR18} states that most existing KGEs are not capable of encoding ontological information.
\cite{iswc21} use ontological information but only for improving the negative samples that KGEs require.

%\begin{figure}[!t]
%	\centering
%	\includegraphics[width=0.7 \columnwidth]{images/idea.pdf}
%	\caption{Approach for combining KGEs with reasoning.\label{fig:idea}}
%\end{figure}

We propose a different approach where we combine the data-driven KGE approach with rule-based reasoning based on ontologies and domain experts and allow them to work in tandem: KGEs are used to extract plausible triples which are used together with the KG for the ontology-based reasoning. At the same time triples inferred by reasoning are used to enhance the prediction of KGEs. This chain continues until no more new information is extracted.
Combining these two approaches is quite challenging for the following reasons:
(i)~The two approaches are completely different: KGE reasoning is inductive (bottom up), making generalizations over input data, while ontology-based rule reasoning is deductive (top-down), reaching to conclusions given general rules. Therefore, it is counter-intuitive on how these two can be combined.
(ii)~KGEs are not capable of generating information out of the box; they are only meant to be able to answer questions, such as 
%finding the subject or object of triples $(?, r, o)$ and $(s, r, ?)$, respectively, or 
whether a triple is true or not. It is thus, not trivial on how we can utilize the KGEs models to extract inferred triples.
(iii)~It is not straightforward on how the two components should communicate as they both require some time to generate information.
%know after how many epochs of training a KGE model, we can extract useful information to pass it to the rule reasoning and vice versa.

We are working towards building a framework which will overcome the above challenges. % allows for KGEs and ontology-based rule reasoning to work in tandem. 
Our framework consists of two main components: a KGE engine and a reasoner. 
%These components iteratively pass newly extracted information to each other. 
The KGE engine produces new triples based on the embeddings and the reasoner produces new triples based on the KG, the ontologies and the rules. The rules can be defined by domain experts or specified by different entailment regimes. 

The novelty of our framework over the previous hybrid approaches is twofold. 
First, it leverages the predictive power of the data-driven KGE approaches while at the same time is able to capture domain expert knowledge via deductive reasoning. This allows to infer information even for sparse regions of a KG. 
Second, it combines KGEs with reasoning in a loosely-coupled way and, thus, does not depend on any specific KGE model or reasoning algorithm: Users can easily input their own KGE or reasoning algorithm and the system will seamlessly work out of the box. 
Thanks to the loosely-coupled combination of KGEs and reasoning, it can achieve better predictive performance.

%Our contribution is a framework that loosely-couples KGEs with ontology-based reasoning and is extensible to any KGE and reasoning algorithms. 
%Our results show that incorporating ontology-based reasoning improves the predictive performance of KG completion.

\section{Preliminaries}
\label{sec:prelim}
%!TEX root = main.tex
Before laying the foundations on KGEs and ontology-based reasoning, let us first formally introduce a knowledge graph.

A knowledge graph $\mathcal{K}$ consists of a set of triples from $\mathcal{E} \times \mathcal{R} \times \mathcal{E}$, where $\mathcal{E}$ is the set of entities and $\mathcal{R}$ the set of relations. A triple $t=(e_i, r, e_j)$ is represented in the graph as two nodes $e_i$ and $e_j$ connected via a directed edge from $e_i$ to $e_j$ with label $r$. We refer to $e_i$ as the subject and to $e_j$ as the object of $t$. For example, the triple $(\texttt{Plato}, \texttt{bornIn}, \texttt{Athens})$ connects the two entities \texttt{Plato} and \texttt{Athens} via the labeled edge \texttt{bornIn}.

\subsection{Knowledge graph embeddings}

\begin{table}
	\caption{Different KGE models.
	$Real()$ is a function that returns the real part of a complex number, $\star$ denotes a circular correlation operation, and $\circ$ denotes ...}\label{tab:kge-models}
\scalebox{0.8}{
\begin{tabular}{ l | c }
	\textbf{KGE model} & \textbf{Scoring function}\\
	\hline
	\hline
	TransE~\cite{transe} &  $-||\mathbf{e}_i + \mathbf{r} - \mathbf{e}_j||_{1~or~2}$\\
%	TransR~\cite{transr} & $-||M_{r} \mathbf{e}_i + \mathbf{r} - M_{r} \mathbf{e}_j ||^{2}_{2}$\\
	DistMult~\cite{distmult} & $\mathbf{e}_i^T diag(\mathbf{r}) \mathbf{e}_j$\\
	ComplEx~\cite{complex} & $Real(\mathbf{e}_i^T diag(\mathbf{r}) \mathbf{e}_j)$\\
	HolE~\cite{HolE} & $\mathbf{r}^T (\mathbf{e}_i \star \mathbf{e}_j)$\\
	RotatE~\cite{rotate} & $-|| \mathbf{e}_i \circ \mathbf{r} - \mathbf{e}_j||^2$ \\
\end{tabular}
}
\end{table}
A KGE maps an entity or a relation to a $d$-dimensional vector ({\em embedding}). 
We denote with $\mathbf{e}_i \in \mathbb{R}^d$ the embedding of an entity $e_i$ and with $\mathbf{r} \in \mathbb{R}^d$ the embedding of a relation $r$.
A KGE model learns a scoring function $f(\mathbf{e}_i, \mathbf{r}, \mathbf{e}_j)$ so that triples $(e_i, r, e_j)$ that exist in $\mathcal{K}$ are generally scored higher than triples that do not exist. The plausibility of a triple $(e_i, r, e_j)$ is then determined by the score output by the function $f(\mathbf{e}_i, \mathbf{r}, \mathbf{e}_j)$.

There have been many proposals on different scoring functions for training KGEs
 resulting in different models. 
Table~\ref{tab:kge-models} lists the most prominent ones.
The objective of training is to optimize the function using some gradient-based algorithm so that it
outputs high scores for the triples that exist in the knowledge graph
and low scores for the triples that do not exist. 
The loss function (e.g,~logistic or pairwise ranking loss) used for the optimization requires both positive and negative triples. 
Negative triples are typically constructed by corrupting an existing triple, i.e.,~replacing its subject or object entities with entities randomly sampled from the graph.

\subsection{Ontology-based reasoning}

\begin{table}
	\caption{A subset of RDFS rules. 
		Relations \texttt{type}, \texttt{domain}, \texttt{range}, \texttt{subProperty}, and, \texttt{subClass} are pre-defined in the RDF/S vocabulary.}
	\label{tab:rdfs}
	\scalebox{0.8}{
	\begin{tabular}{ l | c }
		\textbf{Rule name} & \textbf{Rule (condition $\rightarrow$ consequence)}\\
		\hline
		\hline
		rdfs2 &  $(r, \texttt{domain}, c), (x, r, y) \rightarrow (x, \texttt{type}, c)$\\
		rdfs3 &  $(r, \texttt{range}, c), (x, r, y) \rightarrow (y, \texttt{type}, c)$\\
		rdfs5 &  $(r_1, \texttt{subProperty}, r_2), (r_2, \texttt{subProperty}, r_3) \rightarrow (r_1, \texttt{subProperty}, r_3)$\\
		rdfs7 &  $(r_1, \texttt{subProperty}, r_2), (x, r_1, y) \rightarrow (x, r_2, y)$\\
		rdfs9 &  $(c_1, \texttt{subClass}, c_2), (x, \texttt{type}, c_1) \rightarrow (x, \texttt{type}, c_2)$\\
		rdfs11 &  $(c_1, \texttt{subClass}, c_2), (c_2, \texttt{subClass}, c_3) \rightarrow (c_1, \texttt{subClass}, c_3)$\\
	\end{tabular}
	}
\end{table}

Knowledge graphs are often accompanied by one or more ontologies. 
An ontology gives formal semantics to the knowledge graph by defining the concepts in which entities belong and the relations between these concepts.
For example, the entity \texttt{Plato} belongs to the concept \texttt{Philosopher}, while the entity \texttt{Athens} belongs to the concept \texttt{City}, while concepts \texttt{Philosopher} and \texttt{City} are connected via a relation \texttt{bornIn}.
Using ontologies not only adds semantics to the knowledge graphs but also enables reasoning via rules.
%A rule of the form $head \leftarrow body$ means that if the atom $body$ is satisfied we can infer the atom $head$.

For example, given the rule
\[\forall x,y \in \mathcal{E}: (x,hasCapital, y) \rightarrow (y, locatedIn, x)\]
%\[\forall x,y,z bornIn(x, y), lo \arrowvert locatedIn(y, x)\]
and the triple $(\texttt{Greece}, \texttt{hasCapital}, \texttt{Athens})$ we can infer that \\
$(\texttt{Athens}, \texttt{locatedIn}, \texttt{Greece})$ even if this information is missing from the KG.
By adding more rules and triples, such as
\[\forall x,y,z \in \mathcal{E}: (x, bornIn, y),  (y, locatedIn, z) \rightarrow (x, bornIn, z)\]
and the triple $(\texttt{Plato}, \texttt{bornIn}, \texttt{Athens})$,
we can infer that \\$(\texttt{Plato}, \texttt{bornIn}, \texttt{Greece})$.

Triples such as the latter one are called {\em intensional} (or implicit) triples, while triples that are already present in the initial KG are called {\emph{extensional}} (or explicit).
Rules can be specified by either a domain expert and/or an entailment regime, e.g.,~RDFS or OWL2. Table~\ref{tab:rdfs} shows the rules of a well-defined fragment~\cite{rhordf} of the RDFS entailment regime expressed as Horn clauses.
When an ontology is not provided, such rules can also be discovered via profiling conditional inclusion dependencies~\cite{rdfind}.

\section{Related Work}
\label{sec:related}
%!TEX root = main.tex

Although there is a large body of literature on different KGE models~\cite{complex,distmult,HolE,rescal,transe}  as well as works on scaling KGEs via parallel training~\cite{parallel-kge17,parallel-kge-rainer21}, 
there have been only few works on combining KGEs with logic rules.
One of the first such work is KALE~\cite{kale}, which embeds first-order logic rules in the same mathematical framework of KGEs by devising a new loss function.
In particular, triples are modelled as atomic formulas and are mapped to vectors based on the translation model of TransE~\cite{transe}, while rules are modelled as t-norm fuzzy logics which define the truth value of a complex formula as a composition of the truth values of its constituents.
KALE then tries to minimize the global loss function of both the complex and atomic formulas.
The problem with this approach is that the embedding and reasoning processes are tailored to a specific KGE scoring function which makes the system difficult to adapt to other more efficient and more scalable KGE models. 
pLogicNet~\cite{plogicnet} is based on Markov Logic Network (MLN), which is a rule-based approach utilizing probabilistic graphical models to infer new information and at the same time handle uncertainty. The MLN in pLogicNet is trained with the variational EM algorithm and leverages KGEs to infer missing triples during the inference step (E-step) which are then used in the learning step (M-step) until convergence.
However, by relying on MLNs, pLogicNet inherits the inefficiency and inability to scale to large KGs.
Finally, the authors of~\cite{DWWG-ACL18} impose two simple constraints in KGE models: (i)~vectors of entities should be positive values meaning that we need only positive statements about entities and~(ii)~approximate rules between two relations, e.g., \texttt{bornIn} relation implies \texttt{nationality} relation, which they mine from the data. 
These constraints are then embedded into the score function, e.g.,~of ComplEx, and the KGE model is learned. 
However, this work is coupled with a specific KGE model and the used constraints are very restrictive being unable to incorporate domain knowledge.

\section{Unified Framework}
\label{sec:solution}
%!TEX root = main.tex

\begin{figure}[!t]
	\centering
	\includegraphics[width= \columnwidth]{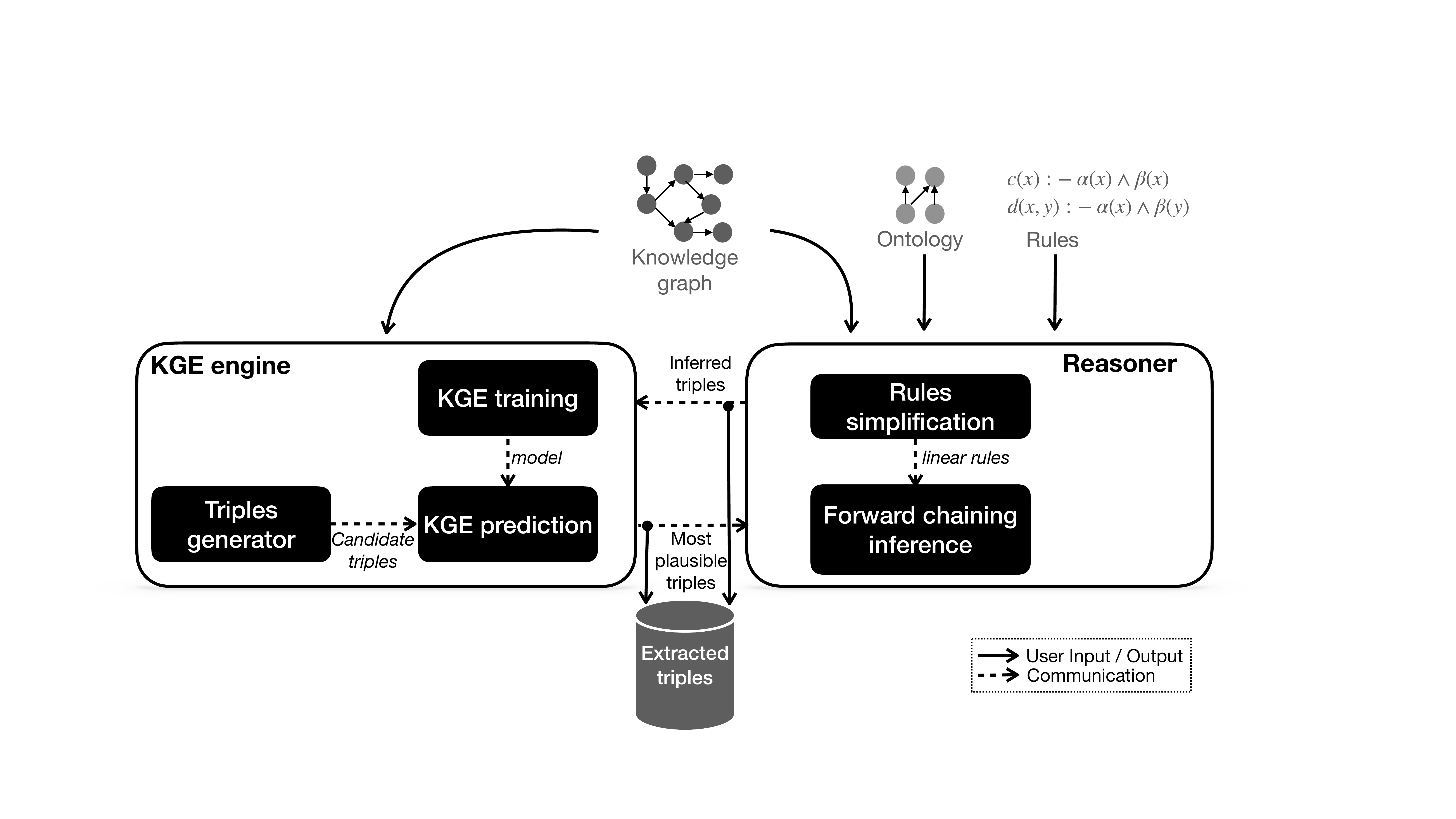}
	\caption{Overview of proposed framework.\label{fig:overview}}
\end{figure}

Our goal is to combine KGEs with ontology-based reasoning in a loosely-coupled way so that we can infer missing information, i.e.,~extract triples that do not exist in a KG. 
Our framework consists of two main components: the KGE engine and the reasoner.
Figure~\ref{fig:overview} shows the internals of the two components and how they interact with each other.
Each component passes its output (i.e.,~inferred tripels) as input to the other component in an iterative manner until no new information can be inferred.

\subsection{KGE engine}
The KGE engine is responsible for the embedding-based learning. 
It receives as input the initial KG together with any triples inferred by the reasoner and trains a KGE model. 
The model is then able to predict with certain probability whether a given test triple, not contained in the training input set, is true or not. 
The challenge we had to overcome is how to generate the test triples.
Ideally, we would like to pass through the KGE prediction the complement of the input KG, i.e.,~all triples that do not exist. 
Although theoretically it is possible to create the complement of the input KG by taking all pairs of nodes that are not directly connected and all possible edge labels, it is infeasible in practice as the number of test triples would be $O(N^2 \times R)$ for a KG with $N$ entities and $R$ relations. 
For this reason, we build the triples generator module inside the KGE engine that is responsible to output a reduced set of candidate triples that have a good chance of being true. 

The triples generator is built in a generic manner to be able to support various methods for generating candidate triples. 
The simplest method is using uniform random sampling to draw from the set of entities and relations. 
This strategy, however, leads to triples that are with high probability not true as connecting randomly entities with random relations creates meaningless triples.
An intuitive strategy is to explore sparse regions of the graph as entities that are densely connected are less probable to have missing true relations.
For this we exploit the cluster coefficient of the entities (nodes in the graph), i.e.,~the fraction of triangles passing through a node w.r.t.~its degree.
We then use the cluster coefficient as a weight to sample entities for each type of relation in KG. 
Finally, the triples generator makes sure that the produced triples are not included in the training set used to build the model.

Once the triples generator module outputs candidates triples, it passes them to the KGE prediction module which outputs whether a triple is true with a certain probability. The KGE engine outputs the triples that are true with a probability above a certain threshold. % (\zoi{0.8} by default).

The goal of the KGE engine is to be independent of any KGE model and triples generation strategy. In other words, users can plug their own implementations. 
To do so, our framework exposes the following primitives:
\vspace{-0.1cm}
\begin{align*}
		\mathtt{model = fit(X)}\\
		\mathtt{Y = generateTriples(X)}\\
		\mathtt{Y' = predict(Y, model)}
\end{align*}

\noindent Function \texttt{fit} receives an array \texttt{X} with the input triples and returns the fitted KGE model. 
Function \texttt{generateTriples} takes as input an array \texttt{X} of triples and outputs another array of candidate triples \texttt{Y} which are not contained in \texttt{X}.
Function \texttt{predict} receives an array of triples \texttt{Y} and a model and outputs the input triples with an extra column that specifies the probability of the triple being true. Our implementation currently includes the following KGE models: TransE, HolE, ComplEx, and DistMult.

\subsection{Inference engine}
The inference engine is responsible to infer new triples given (i)~an initial set of triples (e.g.,~initial KG plus triples output by the KGE engine), (ii)~an ontology, and (iii)~a set of rules. Rules can be user-defined or can be coming from an entailment regime, such as RDFS or OWL2. 
Our framework can use any forward chaining reasoner that exhaustively applies the given rules to the input and inferred triples until a fixpoint is reached, i.e.,~no new triples can be inferred. 
This can be easily achieved by adding the following simple primitive as a layer on top of the reasoner:
\texttt{Y = infer(X, ontology, rules)}. The input is the input triples \texttt{X}, the ontology and the rules.

The type of rules supported strongly depends on the underlying reasoner used by the framework. 
The only requirement is that they should be monotonic, i.e.,~adding new rules or triples never results in removing or contradicting already output results.
In our current implementation, we use Pellet~\cite{pellet} and HermiT\footnote{\url{http://www.hermit-reasoner.com/}} OWL2 DL reasoners.

A problem that may arise with the input rules is that they are redundant, i.e.,~they produce redundant triples. Although this does not change the correctness of the results, it adds a significant overhead on the runtime of the reasoning process. Even the rules included in the RDFS entailment regime are redundant and can lead to a large number of redundant triples~\cite{kaoudi-journal13}. For this reason, we have built a component that simplifies the input rules. 
For example, this module can transform a bilinear (rule rdfs11 Table~\ref{tab:rdfs}) to a linear rule.

\section{Validation}
\label{sec:exp}
%!TEX root = main.tex

We now evaluate our proposal by using both synthetic and real world KGs. The questions we want to answer is 
(i)~how link prediction can be improved by incorporating reasoning with KGEs in a decoupled manner and (ii)~what is the training time overhead that reasoning incurs to the KGE computation.

\header{Datasets} We use datasets one real-world KG that is accompanied with ontology, namely DBPedia, and the popular synthetic KG generator, LUBM.
As the original real KG is too large for KGE models to handle, we use a subset of it: we extracted a well-connected subgraph that contains the 12 most popular relations and information around them.
Table~\ref{tab:datasets} summarizes the characteristics of our datasets.

\begin{table} [t]
\begin{tabular}{ l r r r}
	\hline
	Dataset & Entities & Relations & Triples (total) \\
	\hline
%	YAGO3-10 & 123,182 & 37 & 1,089,040\\
	DBPedia20k & 20,143 & 12 & 120,000 \\
	LUBM-2 & 53,860 & 32 & 268,136 \\
	\hline
\end{tabular}
\caption{Datasets statistics.\label{tab:datasets}}
\end{table}

\header{Baselines} We compare our proposal against the following vanilla KGE models: TransE, DistMult, ComplEx, HolE, and ConvE. In addition, we compare against KALE~\cite{kale} and plogicNet~\cite{plogicnet} which use a different approach to incorporate reasoning in KGEs.
For each dataset and each KGE model, we use the same hyperparameter settings for the vanilla KGEs and our approach. 
For KALE and plogicNet, we use the hyperparameters recommended in their repository. % which lead to the best performance according to the authors.

\header{Hardware and Software}
We ran all our experiments in a machine with 32x2.3 GHz AMD Opteron(tm) processor 6376, 62GB RAM memory, a GPU NVIDIA-SMI 440.33.01, the driver version 440.33.01 and the CUDA version 10.0. We used Python 3.7 and Tensorflow 1.15. 
For the baselines plogicNet and KALE, we used Python 3.7 with PyTorch 1.5 and Java 1.8.0, respectively.

\header{Metrics}
We evaluate the quality of each method by computing the filtered mean reciprocal rank (MRR), and the Hits@1, Hits@3, and Hits@10 for the link prediction task.
These are the standard metrics used to compare different KGE methods. In addition, we measure training time.

\subsection{Model performance}

\begin{table} [t]
	\caption{Quality results among different approaches. 
		Our approach (denoted by an asterisk to the used KGE model) performs better than vanilla KGEs for both LUBM and DBPedia20k.
		It also significantly outperforms the hybrid approches of plogicNet~\cite{plogicnet} and KALE~\cite{kale}. \label{tab:results}}
\scalebox{0.8}{
\begin{tabular}{ l c c c c }
	\multicolumn{5}{c}{\textbf{LUBM}}\\
	\hline
	\textbf{Method} & \textbf{MRR} & \textbf{Hits@1} & \textbf{Hits@3} & \textbf{Hits@10} \\
	\hline
	TransE & 0.28 & 0.28 & 0.26 & 0.31 \\
	TransE* & \underline{0.24} & \underline{0.22} & \underline{0.24} & \underline{0.27}  \\
	\hline
	ComplEx & 0.24 & 0.22 & 0.24 & 0.27 \\
	ComplEx* & \underline{0.32} & \underline{0.27} & \underline{0.32} & \underline{0.41} \\
	\hline
	DistMult & 0.28 & 0.26 & 0.28 & 0.32 \\
	DistMult* & \underline{\textbf{0.35}} & \underline{\textbf{0.32}} & \underline{\textbf{0.36}} & \underline{\textbf{0.40}} \\
	\hline
	HolE & 0.09 & 0.08 & 0.09 & 0.10 \\
	HolE* & \underline{0.27} & \underline{0.25} & \underline{0.27} & \underline{0.29} \\
	\hline
	\hline
	plogicNet & 0.10 & 0.09 & 0.11 & 0.14 \\
	KALE & 0.01 & 0 & 0 & 0.05 \\
	\hline	
	\vspace{0.2cm}
\end{tabular}
}
\scalebox{0.8}{
	\begin{tabular}{ l c c c c }
		\multicolumn{5}{c}{\textbf{DBPedia20k}}\\
		\hline
		\textbf{Method} & \textbf{MRR} & \textbf{Hits@1} & \textbf{Hits@3} & \textbf{Hits@10} \\
		\hline
		TransE & 0.11 & 0.09 & 0.12 & 0.15 \\
		TransE* & \underline{0.14} & \underline{0.10} & \underline{0.15} & \underline{0.20}  \\
		\hline
		ComplEx & 0.25 & 0.18 & 0.27 & 0.38 \\
		ComplEx* & \underline{\textbf{0.29}} & \underline{\textbf{0.22}} & \underline{\textbf{0.33}} & \underline{\textbf{0.43}} \\
		\hline
		DistMult & 0.25 & 0.20 & 0.29 & 0.34 \\
		DistMult* & \underline{0.28} & 0.20 & \underline{0.33} & \underline{0.41} \\
		\hline
		HolE & 0.13 & 0.11 & 0.14 & 0.17 \\
		HolE* & 0.13 & 0.11 & 0.14 & \underline{0.18} \\
		\hline
		\hline
		plogicNet & 0.20 & 0.16 & 0.24 & 0.31 \\
		KALE & 0.18 & 0 & 0 & 0.25 \\
		\hline
	\end{tabular}
}
%\hspace{0.2cm}
%\scalebox{0.75}{
%	\begin{tabular}{ l c c c c }
%		\multicolumn{5}{c}{\textbf{YAGO3-10}}\\
%		\hline
%		\textbf{Method} & \textbf{MRR} & \textbf{Hits@1} & \textbf{Hits@3} & \textbf{Hits@10} \\
%		\hline
%		TransE & 0.49 & 0.38 & 0.55 & 0.67 \\
%		TransE* & 0.44 & 0.33 & 0.50 & 0.64  \\
%		\hline
%		ComplEx & 0.43 & 0.33 & 0.48 & 0.61 \\
%		ComplEx* & 0.38 & 0.28 & 0.43 & 0.58 \\
%		\hline
%		DistMult & 0.46 & 0.37 & 0.52 & 0.63 \\
%		DistMult* & 0.45 & 0.35 & 0.50 & 0.62 \\
%		\hline
%		HolE & 0.45 & 0.37 & 0.51 & 0.61 \\
%		HolE* & 0.44 & 0.35 & 0.50 & 0.60 \\
%		\hline
%		\hline
%		plogicNet & 0.28 & 0.24 & 0.31 & 0.35 \\
%		KALE & 0.32 & 0 & 0 & 0.38 \\
%		\hline
%	\end{tabular}
%}
\end{table}

We first compare the model performance our method achieves compared to vanilla KGE models. Table~\ref{tab:results} shows the results for both datasets.
We observe that for both LUBM and DBPedia20k our approach almost always improves the quality of the model (shown by the underlined times). 
by up to 3x in MRR for the case of HolE in the LUBM dataset. In particular, we improve the MRR, Hits@1, Hits@3, and Hits@10 of HolE~\cite{HolE} by a factor of 3.
The best model for LUBM scored an MRR of 0.35  with DistMult* which includes the DistMult~\cite{distmult} KGE model with our reasoner. For DBpedia20k, ComplEx* scored the best MRR with value of 0.29. As the best KGE model differs for each dataset, it is fundamental to give the opportunity to users to use any KGE model. This is achieved through our decoupled framework.
%The only case where our approach does not perform so well is for the YAGO3-10 dataset. The reason for that is twofold. 
%First, YAGO3-10 has been specifically created for evaluating KGEs: symmetric, antisymmetric and inverse relations have been removed from the training set and added to the test set. This dataset creation process leads to already good performance of KGE models, as it is easier to predict triples with such relations. Adding inferred triples to the model does not give more meaningful information to the KGE algorithm.
%Second, the reasoner generates a lot of inferred triples in the case of YAGO3-10 and thus, makes the training set of a KGE* model much larger. The larger the dataset to train the more epochs are required to reach good performance. 
%Thus, in the last epoch of our experimental evaluation ($4000$th) KGE* lags behind the vanilla KGE model. Still, the gap between the two is very small.

In addition, we observe that our approach significantly outperforms the other hybrid baselines, namely plogicNet~\cite{plogicnet} and KALE~\cite{kale}. 
Our approach achieves up to $3.5 \times$ better MRR performance than plogicNet and $1.5 \times$ better MRR than KALE. This is not only because plogicNet and KALE deeply embed reasoning with KGEs and thus may lose some information but also because our approach can effortlessly use any KGE model and improve over it.

\subsection{Training time}

\begin{figure*}
	\mbox{
		\subfigure[LUBM]{\includegraphics[width = 0.32 \linewidth] {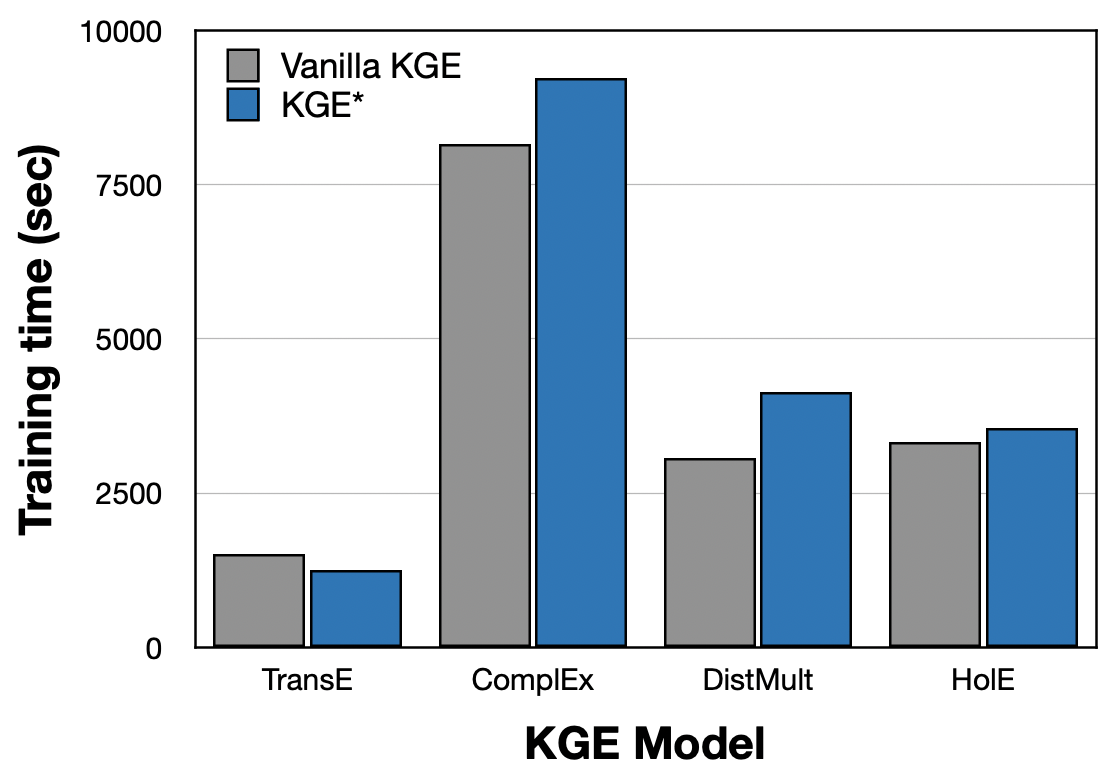}\label{subfig:time-lubm}}
		\subfigure[DBPedia20k]{\includegraphics[width = 0.32 \linewidth] {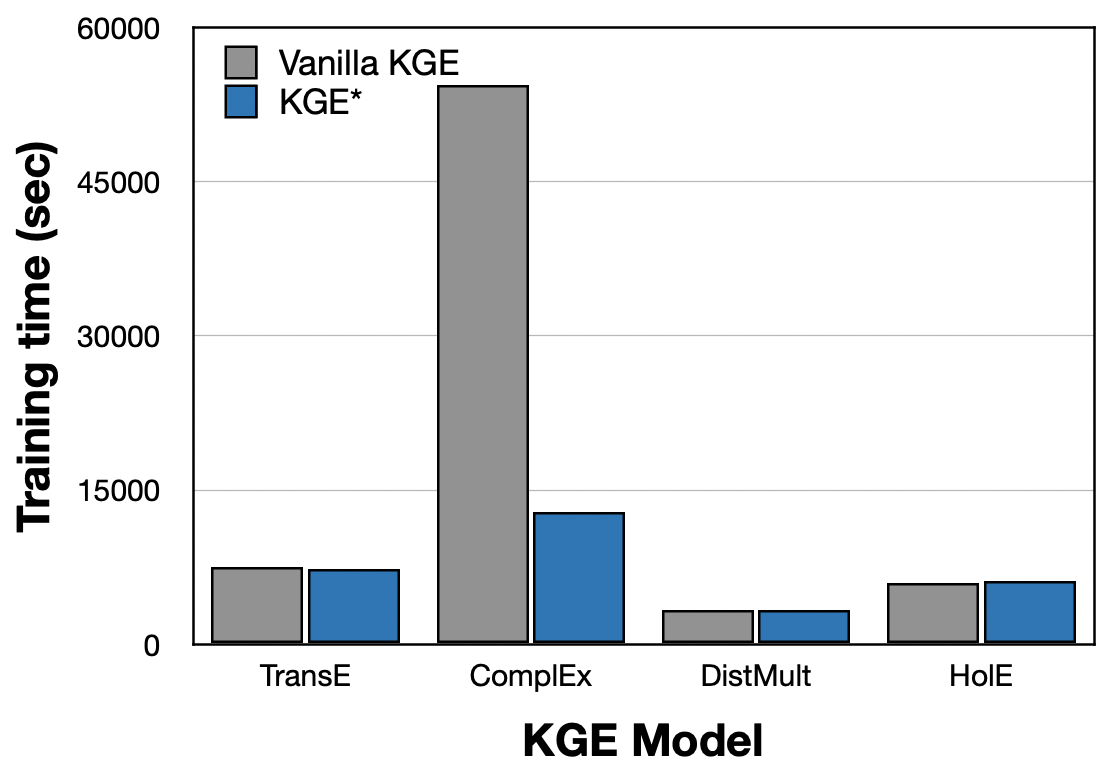}\label{subfig:time-dbpedia}}
%		\subfigure[YAGO3-10]{\includegraphics[width = 0.3 \linewidth] {images/training_time-yago.png}}
		\subfigure[vs. KALE and plogicNet]{\includegraphics[width = 0.3 \linewidth] {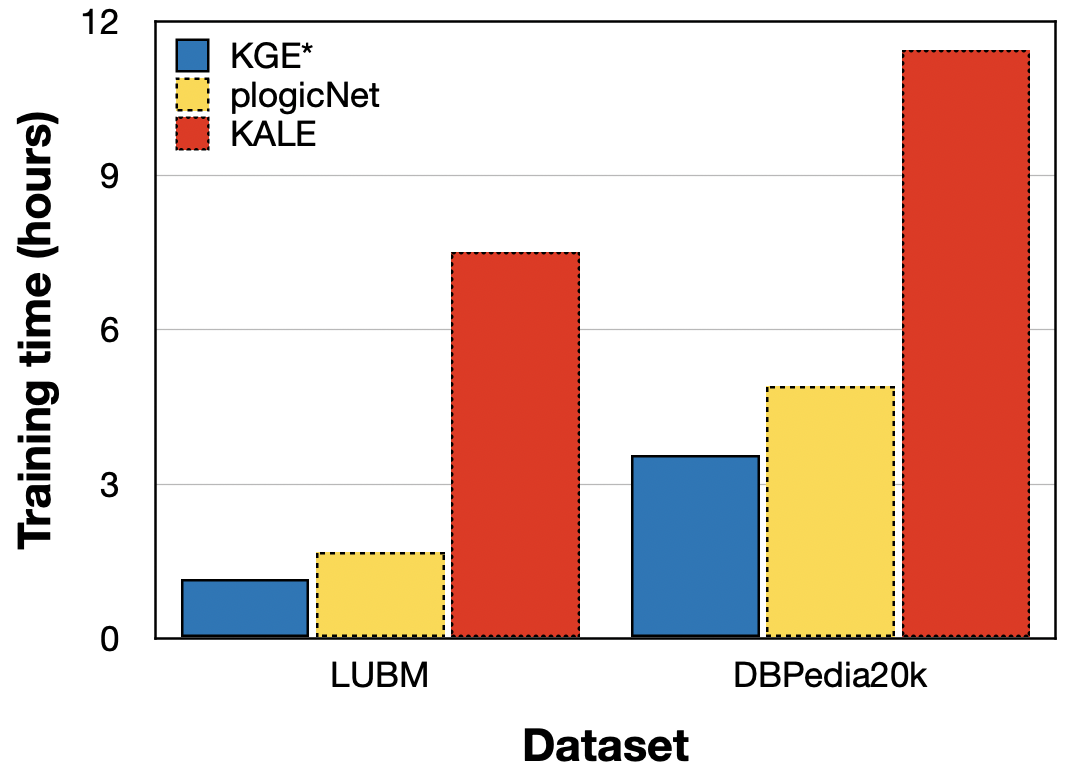}\label{subfig:time-baselines}}
	}
	\caption{(a) and (b): Training times for vanilla KGE and improved with reasoning KGE (KGE*) for different datasets: KGE* keeps the reasoning overhead low for most of the cases. Adding reasoning may result to faster convergence and thus lower training times (see e.g.,~ComplEx for DBpedia). (c): KGE* is much faster than plogicNet and KALE for both LUBM and DBPedia20k datasets.
		%However, there is still room for improvement as shown for the YAGO3-10 dataset. 
		\label{fig:time}}
\end{figure*}

%\begin{figure}
%		\includegraphics[width = 0.8 \columnwidth] {images/training_time_baselines.png}
%	\caption{KGE* is much faster than plogicNet and KALE for both LUBM and DBPedia20k datasets.\label{fig:time}}
%\end{figure}

We now evaluate what is the overhead of adding the reasoning process when training KGEs. Figure~\ref{fig:time} shows the training time in hours for (i)~different vanilla KGE models compared to their counterpart KGE* and~(ii)~the baselines compared to the KGE* model that yields the best prediction accuracy.
Figure~\ref{subfig:time-lubm}, we observe that the overhead posed by the reasoning engine is very small for the LUBM dataset (up to $34\%$ for DistMult), while for TransE the inferred triples resulting from the inference lead to faster convergence of the algorithm.  Similar results we observe for the DBpedia20k dataset in Figure~\ref{subfig:time-dbpedia}.

Finally, Figure~\ref{subfig:time-baselines} shows the difference in training time between KGE* and the hybrid baselines KALE and plogicNet. For both datasets, KGE* is faster to converge than the baselines that need significant much more time.

\section{Conclusion}
\label{sec:conclusion}
%!TEX root = main.tex

We proposed a hybrid framework that loosely decouples KGEs with ontology-based reasoning. 
In contrast to state-of-the-art approaches that rely solely on the information available in the input knowledge graph, we can incorporate domain expertise in the KG completion task which leads to increased accuracy.
The loose decoupling also allows users to plug any KGE or reasoning algorithm they like allowing for further optimizations. Our results showed that our proposed framework can improve the accuracy of prediction by up to 3x while keeping the training time low.

\bibliographystyle{ACM-Reference-Format}
\bibliography{references}

% \newpage
% \input{sections/appendix}

\end{document}